# Advanced spectral clustering for heterogeneous data in credit risk monitoring systems

Lu Han[1]   Mengyan Li[1]   Jiping Qiang[2]   Zhi Su[3, *]

[1] School of Management Science and Engineering, *Central University of Finance and Economics,* Beijing 100081, China

[2] School of Information Engineering, *Yangzhou University*, Yangzhou 225127, China

[3] School of Statistics and Mathematics, *Central University of Finance and Economics,* Beijing 100081, China

[*] *Corresponding author E-mail:* zhisu@cufe.edu.cn

**Abstract**: Heterogeneous data, which encompass both numerical financial variables and textual records, present substantial challenges for credit monitoring. To address this issue, we propose Advanced Spectral Clustering (ASC), a method that integrates financial and textual similarities through an optimized weight parameter and selects eigenvectors using a novel eigenvalue-silhouette optimization approach. Evaluated on a dataset comprising 1,428 small and medium-sized enterprises (SMEs), ASC achieves a Silhouette score that is 18% higher than that of a single-type data baseline method. Furthermore, the resulting clusters offer actionable insights; for instance, 51% of low-risk firms are found to include the term 'social recruitment' in their textual records. The robustness of ASC is confirmed across multiple clustering algorithms, including k-means, k-medians, and k-medoids, with ΔIntra/Inter < 0.13 and ΔSilhouette Coefficient < 0.02. By bridging spectral clustering theory with heterogeneous data applications, ASC enables the identification of meaningful clusters, such as recruitment-focused SMEs exhibiting a 30% lower default risk, thereby supporting more targeted and effective credit interventions.

**Keywords**: Advanced spectral clustering; Heterogeneous data; Group portrait; Credit monitoring

**Acknowledgment:** The work was supported by the National Natural Science Foundation of China (No. 72101279, No. 72173145, No. 72134002), Visiting Scholar Grant Program of China Scholarship Council for Han (No. 202406490006) and the Fundamental Research Funds for the Central Universities (No. JYXZ2407).

**Authors' Email**:

Lu Han: hanluivy@126.com

Mengyan Li: kristina5216@163.com

Jipeng Qiang: jpqiang@yzu.edu.cn

Zhi Su: zhisu@cufe.edu.cn

# Advanced spectral clustering for heterogeneous data in credit risk monitoring systems

**Abstract**: Heterogeneous data, which encompass both numerical financial variables and textual records, present substantial challenges for credit monitoring. To address this issue, we propose Advanced Spectral Clustering (ASC), a method that integrates financial and textual similarities through an optimized weight parameter and selects eigenvectors using a novel eigenvalue-silhouette optimization approach. Evaluated on a dataset comprising 1,428 small and medium-sized enterprises (SMEs), ASC achieves a Silhouette score that is 18% higher than that of a single-type data baseline method. Furthermore, the resulting clusters offer actionable insights; for instance, 51% of low-risk firms are found to include the term 'social recruitment' in their textual records. The robustness of ASC is confirmed across multiple clustering algorithms, including k-means, k-medians, and k-medoids, with $\Delta$Intra/Inter < 0.13 and $\Delta$Silhouette Coefficient < 0.02. By bridging spectral clustering theory with heterogeneous data applications, ASC enables the identification of meaningful clusters, such as recruitment-focused SMEs exhibiting a 30% lower default risk, thereby supporting more targeted and effective credit interventions.
**Keywords**: Advanced spectral clustering; Heterogeneous data; Group portrait; Credit monitoring

## 1. Introduction

Credit monitoring, which differs fundamentally from credit evaluation, emphasizes the continuous tracking of borrower behavior post-loan disbursement rather than pre-loan financial assessments. This task poses unique challenges for small and medium-sized enterprises (SMEs), given their irregular financial reporting practices and heavy reliance on manually processed textual audit records (Jiang et al., 2023; Lee et al., 2024). While existing studies predominantly focus on numerical financial data, such data often suffer from issues related to timeliness and estimation biases (Kozodoi et al., 2025; Mohammadnejad-Daryani et al., 2024; Nallakaruppan et al., 2024). Moreover, the manual processing of textual records introduces inefficiencies and subjective interpretations. Consequently, there is a pressing need to develop models capable of effectively handling heterogeneous data.

Credit monitoring is always an unsupervised learning process that focuses on grouping loaned SMEs without distinguishing between acceptance or rejection. It primarily uses clustering methods to create group profiles, enabling commercial banks to provide differentiated services such as credit extensions or collections. Most clustering methods analyze numerical data through distance calculations, while textual data clustering has grown in importance due to the increasing volume of available data (Costa & Ortale, 2025). However, few studies currently address heterogeneous data clustering, making it a new challenge in data analysis to develop methods that integrate both types effectively.

Spectral clustering a kind of graph-based methods, uses Laplacian matrix decomposition to identify non-convex cluster structures, overcoming the limitations of centroid-based methods (Naseri et al., 2025; Wang et al., 2024). Its eigenvector-based dimensionality reduction makes it suitable for high-noise, high-dimensional, heterogeneous data (Wen et al., 2021).

This study introduces the Advanced Spectral Clustering (ASC) method, which integrates financial statements and loan audit texts to address clustering challenges dealing with heterogeneous data. The primary objectives and contributions of this study are as follows:

(1) To propose an effective unsupervised learning approach for heterogeneous data, addressing key challenges in data fusion analysis for such datasets.

(2) To overcome the difficulty in selecting the number of clusters in spectral clustering, we develop an optimization function that balances eigenvalues with clustering outcomes, thereby achieving superior performance.

(3) To construct group profiles applicable to credit monitoring systems used by most financial institutions.

The rest of this paper is structured as follows. Section 2 reviews the related work, focusing on two key areas: clustering with heterogeneous data and spectral clustering. Section 3 presents the Advanced Spectral Clustering method in detail. Section 4 demonstrates and analyzes the experimental results. Section 5 discusses the applicability of our approach. Section 6 summarizes the conclusions and outlines future research.

## 2. Related work

The main challenge in heterogeneous data clustering is effectively integrating information from numeric, categorical, and ordinal attributes, as well as text and images. Recent advances have focused on calculating similarity matrix for heterogeneous data clustering (Gupta, Thakar, & Tokekar, 2025). Kuo et al. (2024) integrate genetic algorithms with sine-cosine algorithms to optimize attribute weights and initial centroids, incorporating Gaussian, Levy, and single-point mutations to avoid local optima. Wang & Mi (2025) propose an Intuitive-K-prototypes algorithm that enhances prototype representation and defines intuitionistic distribution centroids for categorical attributes. For textual data, Roy & Basu (2022) employ spectral clustering to reduce corpus size and calculate similarity based on both content and shared neighbours using cosine similarity. Liu et al. (2024) utilize BM25 similarity and TR scores for text summarization to minimize redundancy in long documents. Kar et al. (2025) develop the EDMIX metric, which uses Boltzmann entropy to capture categorical attribute distributions and adaptively balances numerical and categorical contributions. Zhang et al. (2025) propose a heterogeneous attribute reconstruction and representation learning paradigm that addresses implicit associations between attributes through uniform metric learning.

Spectral clustering, a graph theory-based method, has received significant attention in machine learning and data mining. It treats data points as graph nodes, captures relationships via a similarity matrix, and maps data to a low-dimensional space using Laplacian matrix decomposition for clustering. Relevant studies include Bruneau & Otjacques (2018), Matsuda, et al. (2020) and Gao et al. (2024). Its ability to handle complex nonlinear data structures makes it suitable for stream learning and multi-view data clustering (Wang et al., 2024). Due to its high performance and simplicity, spectral clustering is widely applied in image segmentation and text clustering (Zhu et al., 2019; Zhou et al., 2022).

Spectral clustering involves two key challenges: similarity matrix construction and computational efficiency. For matrix construction, researchers have proposed various optimization strategies. Houthuys, et al. (2018) formulates the clustering problem using weighted kernel canonical correlation analysis within a primal-dual optimization framework based on Least Squares Support Vector Machines. Favati et al. (2020) investigated sparsity distance calculation using minimum spanning trees to optimize Gaussian kernel values. Zhu et al. (2022) introduced a dynamic hypergraph learning framework for joint optimization of hypergraph structure and feature weights in sparse feature selection. Jia et al. (2023) developed a global and local structure preserving model within the non-negative matrix decomposition framework. Naseri et al. (2025) proposed a subspace distance-based similarity metric that addresses overlapping subspaces through local neighbourhood analysis. Additionally, Zhu et al. (2025) improved user dissimilarity measurement using Wasserstein distance in spectral clustering. Despite these advances, no unified standard exists for constructing similarity matrices across application domains, leaving room for further exploration. The second challenge is computational efficiency, as Laplacian matrix decomposition typically has $O(n^2)$ complexity. Studies addressing this issue include He et al. (2025), Wang et al. (2024) and Gao et al. (2024). In our research, given the sample size, we focus on similarity matrix construction.

The reviewed literature highlights the persistent challenges in heterogeneous data clustering, particularly in effectively integrating heterogeneous data (numeric, categorical, textual) and the strong reliance of spectral clustering performance on similarity matrix construction. Despite notable advancements, a significant gap remains. There is still a lack of robust, interpretable, and high-performing spectral clustering frameworks capable of seamlessly integrating fundamentally different data modalities, such as structured financial data and unstructured text, while simultaneously optimizing the spectral embedding process.

## 3. Advanced Spectral Clustering

In this section, we propose an advanced spectral clustering method (ASC) for heterogeneous data. The method consists of three primary steps: (1) constructing the similarity matrix, (2) generating the Laplacian matrix and selecting k eigenvalues, and (3) applying eigenvector-based clustering. The detailed procedures are described below.

### *3.1 Constructing similarity matrix*

Assuming that the data set mainly consists of numerical data and textual data. Due to significant differences in similarity calculation between these two types of data, we must construct a new distance function to calculate the similarity.

For the numerical data, Mahalanobis distance is often used to dealing with data points that is highly dependent on the distribution. Let $\Sigma$ be the $p \times p$ covariance matrix of the data set. The $(i, j)$ th entry of the covariance matrix is equal to the covariance between the dimensions i and j. Then, the Mahalanobis distance between two p-dimensional data points can be calculated as equation (1):

$$Maha(x_i, x_j) = \sqrt{(x_i - x_j)\sum\nolimits^{-1}(x_i - x_j)^T} \qquad (1)$$

Then the similarity function can be calculated as equation (2):

$$Sim(x_i, x_j) = \frac{\max(Maha(x_i, x_j))}{Maha(x_i, x_j)} \qquad (2)$$

For textual data, it can be regarded as quantitative multidimensional data when treated as a bag of words. The frequency of each word serves as a quantitative attribute, and the base lexicon represents the full set of attributes. The most common method to measure similarity between different texts is by using the normalized Cosine measure. Let $t_i$ and $t_j$ be two text records on a lexicon of size $q$. The inverse document frequency id then can be used for normalization as equation (3):

$$id_i = \log(n / n_i) \qquad (3)$$

Where, n is the total number of words, $n_i$ is the number of words in $t_i$. Then we can normalize frequency $h(t_i)$ for the ith text record using a damping function $f(t_i) = \log(t_i)$, which can be calculated as equation (4):

$$h(t_i) = f(t_i) \cdot id_i = \log(t_i) \cdot \log(n / n_i) \tag{4}$$

Then, the normalized Consine measure can be defined as equation (5), and it is also the similarity function for two textual data.

$$Sim(t_i, t_j) = \frac{\sum_{i=1, j=1}^{q} h(t_i) \cdot h(t_j)}{\sqrt{\sum_{i=1}^{q} h(t_i)^2} \cdot \sqrt{\sum_{j=1}^{q} h(t_j)^2}} \tag{5}$$

It is straightforward to extend the approach to heterogeneous data by incorporating the weights of numerical similarity and textual similarity. The overall similarity between two samples can then be defined as in equation (6):

$$Sim_{i,j} = \lambda \cdot Sim(x_i, x_j) + (1 - \lambda) \cdot Sim(t_i, t_j) \tag{6}$$

If we know that some samples must be linked to certain samples, while others must not be linked to specific samples, we can define two sets: the must-link set denoted as M, and the cannot-link set denoted as C. Then we can then use the optimization function in equation (7) to find the best parameter $\lambda$.

$$\begin{aligned}
&\max \lambda \cdot (1 - \lambda) \\
&s.t. \\
&\lambda \cdot Sim(x_i, x_j) + (1 - \lambda) \cdot Sim(t_i, t_j) \geq Sim(x_i, x_k) \\
&\lambda \cdot Sim(x_i, x_j) + (1 - \lambda) \cdot Sim(t_i, t_j) \geq Sim(x_j, x_k) \\
&\lambda \geq 0 \\
&\lambda \leq 1 \\
&x_i, x_j \in M \\
&x_k \in C
\end{aligned} \tag{7}$$

If we obtain the parameter $\lambda$, we can construct a graph such that the local clustering structure of the graph is preserved by embedding data points into multidimensional space. Let $G = (N, E)$ denote the undirected graph with node set $N$ and edge set $E$. A symmetric weight matrix $W$ defines the corresponding node similarity based on the specific choice of neighbourhood transformation. Consequently, all entries in the matrix are assumed to be non-negative, with higher values indicating greater similarity.

## 3.2 Generating the Laplacian matrix and selecting k eigenvalues

When we construct the graph based on similarity matrix $W$ with data points, we can then calculate the degree matrix $D$. Here $D$ is a diagonal matrix, that the elements in $D$ is on the diagonal representing the degree $d_i$ of the ith node, which can be calculated by equation (8), and the other elements are all zeros, as the work of Chifu et al. (2015).

$$d_i = \sum_{}^{n} w_{ij} \tag{8}$$

The Laplacian matrix can be defined as equation (9), and it is positive semi-definite with non-negative eigenvalues.

$$\begin{aligned} L &= D - W \\ L_{sym} &= D^{-1/2} L D^{-1/2} \end{aligned} \tag{9}$$

Since $D^{-1} L \bar{y} = \lambda \bar{y}$, then finding the optimal k-dimensional embedding can be achieved by determining eigenvectors of $D^{-1}L$ with successively increasing eigenvalues. After discarding the first trivial eigenvector $e_1$ with $\lambda_1 = 0$, this will result in a set of $k$ eigenvectors $e_2, e_3, \cdots, e_k$ with corresponding eigenvalues $\lambda_2 \leq \lambda_3 \leq \cdots \lambda_k$. The ith component of the jth eigenvector represents the jth coordinate of the ith data point.

So, it will need to choose $k$ eigenvectors to embed the data, and $k$ is the parameter that has the most significant impact on the results. Usually, the eigenvalue difference is used to determine the data of $k$, and a sudden increase in eigenvalues will be taken as a signal for selecting $k$. Of course, this approach only considers the value of the feature, without considering the clustering results. Here we propose an optimization function balance the difference between eigenvalues and Silhouette Score for selecting $k$ more reasonable, which is calculated as equation (10):

$$\min_{k} \max(\Delta e_i + \frac{|\bar{A}_i - \bar{C}_i|}{\max\{\bar{A}_i, \bar{C}_i\}}) \tag{10}$$

where $\Delta e_i$ is the differences between eigenvalues, and $\bar{A}_i$ is the average distance in ith cluster, and $\bar{C}_i$ is the minimum average distance from the ith cluster centre to other cluster centres.

When selecting the best $k$, then we can create an $n \times k$ matrix, corresponding to a new k-dimensional representation of each of the $n$ data points. Then clustering algorithm, such as k-means, can be applied to the transformed matrix to find the clusters. The whole process of the method is as follow:

| |
|---|
| *Algorithm 1*. **Advanced Spectral Clustering** |
| **Inputs:** $n \times m$ data set {xi}, which has p-dimensional numerical data, q-dimensional textual data |
| **Outputs:** clusters |
| **Step 1:** Calculate similarity on numerical data using Eq. (2) |
| **Step 2:** Calculate similarity on textual data using Eq. (5) |
| **Step 3:** Solve the Optimization (7) for choose $\lambda$ |
| **Step 4:** Construct the optimal similarity matrix W |
| **Step 5:** Calculate the Laplacian matrix using Eq. (9) |
| **Step 6:** Decompose eigenvalue of the Laplacian matrix |
| **Step 7:** Construct a new transformation matrix using the eigenvector corresponding to the smallest i-th eigenvalue |
| **Step8:** Apply K-means clustering on the new $n \times k$ matrix, and output the clusters |
| **Step9:** Optimize k using Eq. (10), and return to step 7, until there is no k that can be optimized. |

## 4. Data and Results

### *4.1 Data*

In this section, we demonstrate the application of advanced spectral clustering through experiments on a loan audit dataset. The data consist of loan records for **1,866 small and medium-sized enterprises (SMEs)** from a city commercial bank in China in 2020, with 1,428 SMEs approved and 438 rejected. The dataset is available at the following link:
https://www.researchgate.net/publication/390694764_Credit_investigation_of_SMEs.

Using financial data reported by these enterprises, we selected **5 financial ratios** as numerical variables based on the classic credit evaluation Z-score model: Current Assets/Total Assets (Var1), Retained Earnings/Total Assets (Var2), Net Profit/Total Assets (Var3), Equity/Total Liabilities (Var4), and Operating Income/Total Assets (Var5). Basic statistics for these variables are presented in Table 1.

**Tabel 1.** Basic statistics of five variables of the sample

| Variable | Mean(%) | Standard Deviation | Coefficient of Variation |
|---|---|---|---|
| **Var1** | 21.34 | 19.65 | 92.08 |
| Var2 | 17.28 | 7.59 | 43.92 |
| **Var3** | 15.38 | 12.37 | 80.43 |
| Var4 | 64.53 | 20.63 | 31.97 |
| Var5 | 82.36 | 22.57 | 27.40 |

Since credit officers conduct quarterly surveys on these enterprises, the survey texts of the 1,428 SMEs serve as the textual data for our experiment. Key terms (e.g., 'inventory', 'recruitment') were extracted using Baidu NLP, resulting in a **236-word lexicon** after stop-word removal. Some sample texts are presented in Table 2, while part of the frequency and cumulative frequency of the entities are shown in Figure 1. For more details, refer to Han et al. (2023).

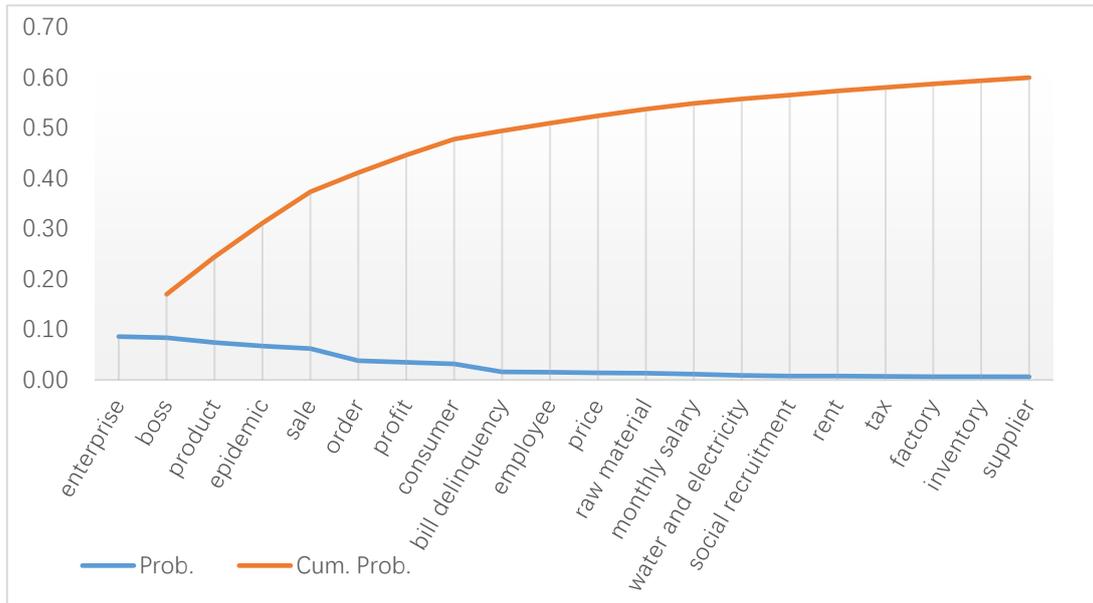

**Figure 1.** Part of frequency and cumulative frequency of entities

The experimental workflow is illustrated in Figure 2, where all experiments are conducted using MATLAB 2024b. As shown in the figure, the upper section depicts the process of ASC, while the lower section outlines the robustness experiments and comparisons with baseline models. The experimental results are organized into three parts: (1) a comprehensive explanation of the ASC procedure, (2) robustness testing of the proposed algorithm through single-type data and classical clustering methods, and (3) comparative analysis between ASC and baseline models to derive general conclusions.

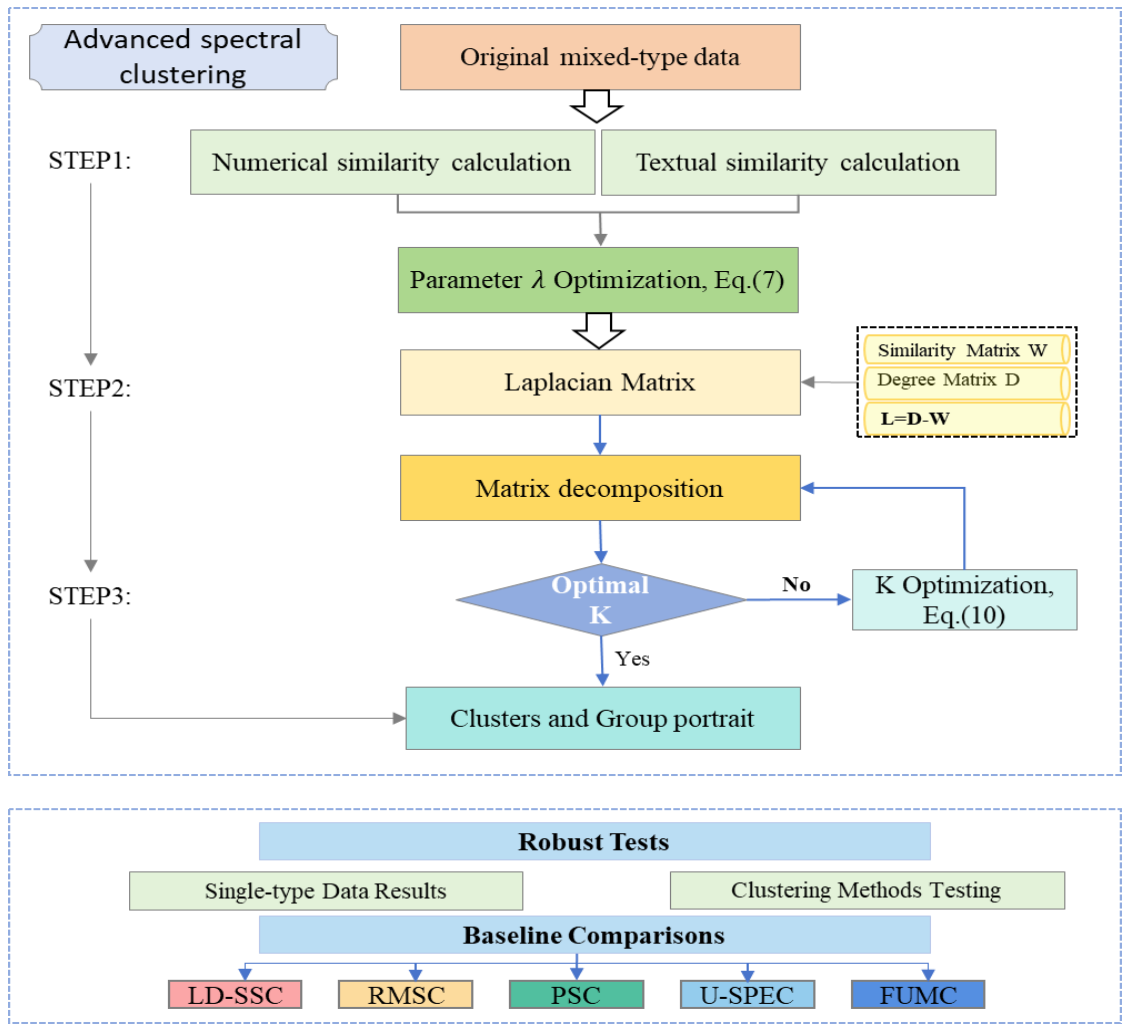

**Figure 2.** Workflow of the experiments

**Table 2.** Samples of the texts and labelled words

| Original texts | Translations | Labeled words |
|---|---|---|
| 该企业在 5 月 27 日支付 36.72 万元购买济南恒鹏钢铁公司钢材。企业有生产员工 11 人。最近销量比较好。未联系上老板。 | On May 27th, this enterprise paid 367,200 yuan to purchase steel from Jinan Hengpeng Steel Company. The enterprise has 11 production staff. The sales have been good recently. The boss could not be contacted. | enterprise; pay; purchase; steel; Jinan Hengpeng steel company; enterprise; production; staff; sale; good; boss; not be contacted. |
| 该企业无水电欠费记录，销售经理月收入大概 8000 元，最近快两个月周末都在加班。 | This enterprise has no record of unpaid water or electricity bills. The sales manager's monthly income is approximately 8,000 yuan. Recently, the staff have been working overtime on weekends for nearly two months. | enterprise; no record of unpaid; water or electricity; bill; sales manager; monthly income; work overtime on weekends. |
| 老板去北京学习了，没有见到本人。有 4 个人在公司干活，今天上午前后来了五六个客户。 | The boss went to Beijing for study and couldn't meet with us in person. There are four people working in the company. Around 6 or 7 clients came to the company this morning. | boss; go to Beijing; study; not meet; people; work; company; clients; come. |
| 园区登记有 1 个月的水费拖欠。公司有 20 人左右在办公，老板不在公司，人力部门负责人接待，正在进行社会招聘，司机月薪 4000，有三险一金及加班补贴，技术人员月薪 7000，有三险一金，招聘 10 人。 | There is a 1-month overdue water bill registered in the park. There are about 20 people working in the company. The boss is not in the company. The head of the human resources department is handling the situation. They are currently conducting social recruitment. The driver's monthly salary is 4000 yuan. They have three insurances, one housing fund and overtime subsidies. The technician's monthly salary is 7000 yuan. They have three insurances and one housing fund. They are recruiting 10 people. | overdue; water bill; people; work; company; boss; be not in; head of the human resources; department; social recruitment; monthly salary; driver; three insurances; one housing fund; overtime subsideies; technician; montly salary; three insurances; one housing fund; recruit. |

| 老板说这个月订单扩大了一倍，但是因为疫情，钢材原材料价格涨了快一半，目前产品没有提价，计划后面原材料价格再涨价就提高产品价格，毕竟现在利润太低了。 | The boss said that this month's orders have doubled, but due to the pandemic, the price of steel raw materials has increased by nearly half. Currently, the products have not been raised in price. The plan is to increase the product prices if the raw material prices rise again later. After all, the profit is currently too low. | boss; say; order; be doubled; epidemic; price; steel; raw material; rise; product; rise in price; plan; increase; product price; raw material; rise; profit; low. |

*4.2 Advanced spectral clustering*

In the experiment, we define the 1,428 approved SMEs as the must-link set and the 438 rejected SMEs as the cannot-link set.

We utilize **5 financial ratios** to compute the Mahalanobis distance for the 1,866 SMEs and select the maximum value to construct the similarity function for each SME based on Equation (2).

We select the top **20 entities** with the highest frequency, along with their associated behaviors and descriptions, to construct a basic matrix for the texts. Subsequently, we apply Equation (5) to calculate the similarity scores for each of the 1,428 SMEs in the must-link set.

Using the must-link set and the cannot-link set, we substitute the similarity values of each data point and optimize the parameters $\lambda$ at intervals of 0.05. The optimal value is 0.65. The average similarities of the must-link set and the cannot-link set under different parameter settings are presented in Table 3.

**Table 3.** The average similarity of must-link set and cannot-link set with different $\lambda$

| $\lambda$ | Average Similarity (Must-link set) | Average Similarity (Cannot-link set) |
|---|---|---|
| 0 | 0.57 | 5.61 |
| 0.05 | 0.62 | 5.87 |
| 0.10 | 0.63 | 5.66 |
| 0.15 | 0.61 | 5.73 |
| 0.20 | 0.64 | 5.74 |
| 0.25 | 0.65 | 5.83 |
| 0.30 | 0.63 | 5.92 |
| 0.35 | 0.68 | 6.12 |
| 0.40 | 0.64 | 6.11 |
| 0.45 | 0.65 | 6.18 |
| 0.50 | 0.63 | 6.19 |
| 0.55 | 0.68 | 6.24 |
| 0.60 | 0.69 | 6.34 |
| **0.65** | **0.72** | **6.88** |
| 0.70 | 0.69 | 6.84 |
| 0.75 | 0.67 | 6.73 |
| 0.80 | 0.61 | 6.64 |
| 0.85 | 0.58 | 6.52 |
| 0.90 | 0.52 | 6.45 |
| 0.95 | 0.51 | 6.17 |
| 1 | 0.47 | 5.96 |

Furthermore, we compute the Laplacian matrix and perform spectral decomposition. After removing eigenvalues that are approximately zero, we select the 20 smallest eigenvalues to calculate the maximum increment between consecutive eigenvalues and determine the optimal $k$ value. These eigenvalues and the corresponding increments are illustrated in Figure 3.

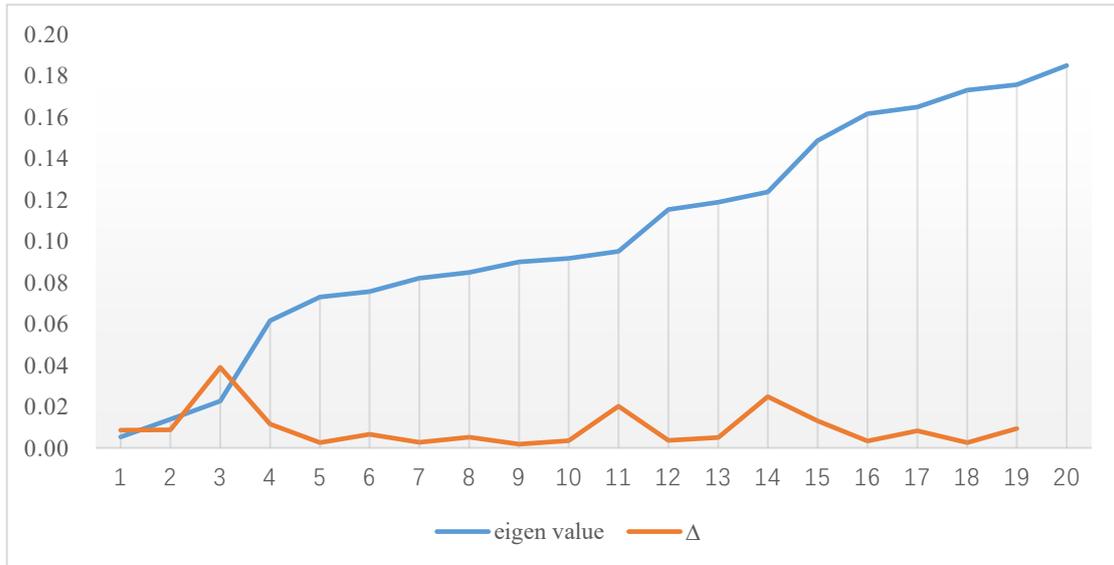

**Figure 3.** The minimum k eigenvalues of Laplacian matrix and its Δ

From Figure 3, it can be observed that there are three $k$ values corresponding to a rapid increase in eigenvalues: 3, 11, and 14. Furthermore, we perform k-means clustering on these selected $k$ values and determine the optimal $k$ value based on the Silhouette Score of the clustering results. The Silhouette Scores are presented in Figure 4. As shown in Figure 4, the optimal $k$ value is determined to be 3.

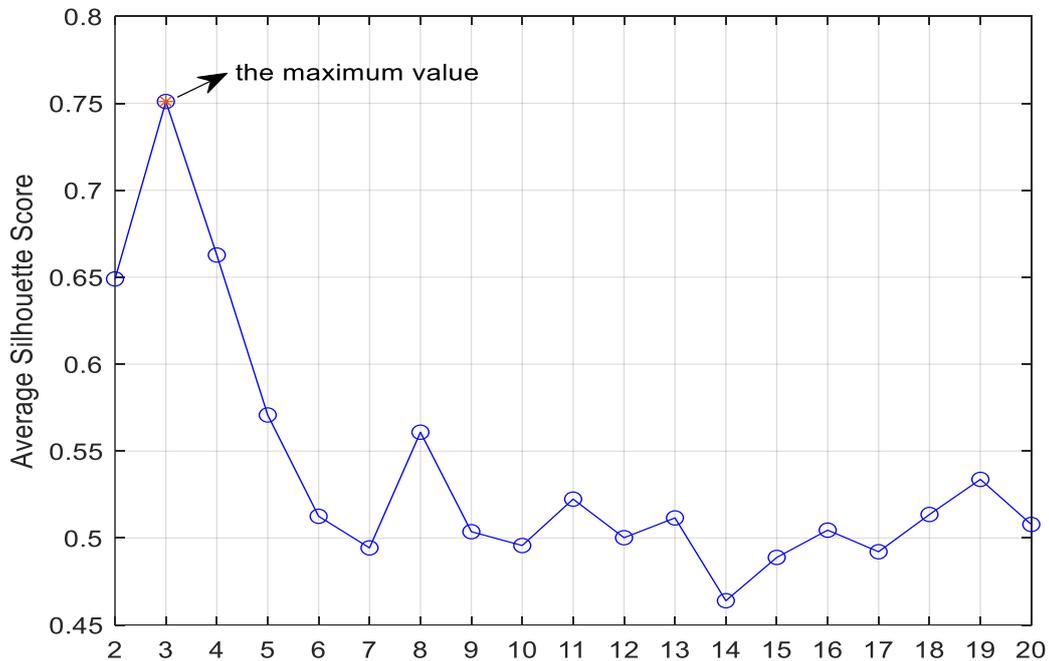

**Figure 4.** Average Silhouette Score under different k

Using the optimal $k$ value of 3, we perform k-means clustering on the combinations of the three eigenvectors of the Laplacian matrix to obtain labels for each data point. Furthermore, we select the two variables with the highest coefficient of variation (CV) as the basic dimensions to visualize the clusters, as illustrated in Figure

5.

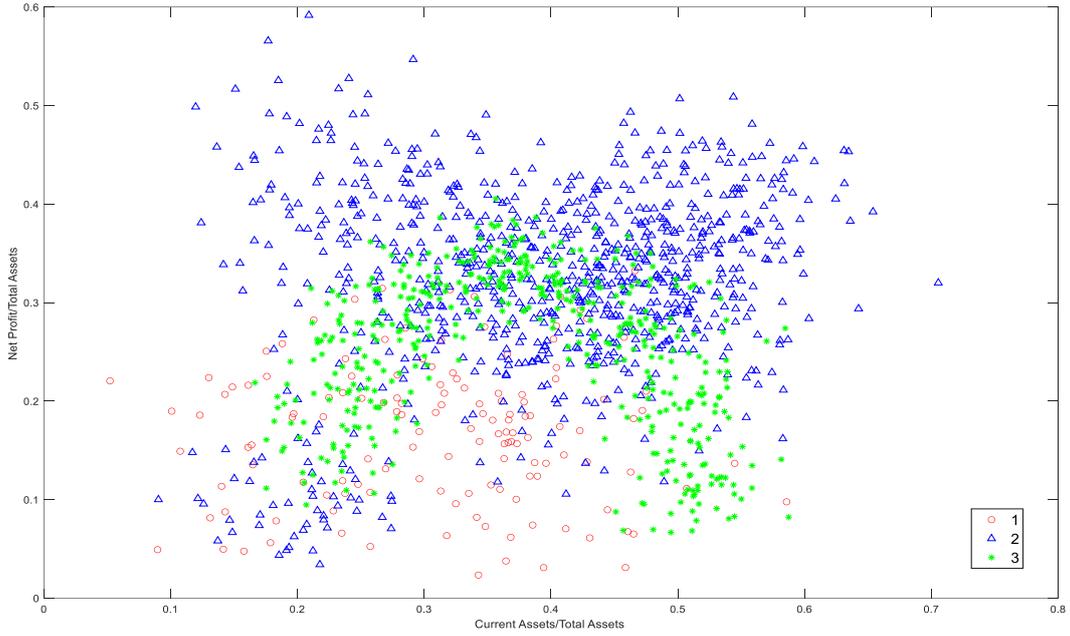

**Figure 5.** Samples in different clusters under the highest CV dimension

From Figure 5, it can be observed that in the two major feature dimensions—current assets/total assets and net profit/total assets—the data distribution is not well structured. However, our clustering algorithm captures the contour morphology of the clusters, even when they are not fully connected. Moreover, the cluster labelled as green exhibits a relatively higher degree of connectivity.

We further analyze the word frequency ratio of the textual data within these three clusters, which is calculated as follows:

$$ratio = \frac{\sum \text{Number of the entity in the category}}{\sum \text{The total number of the entity}} \tag{11}$$

The results are presented in Table 4. As shown in Table 4, there are minimal differences in the distribution of high-frequency words across the three categories; however, these differences are not statistically significant. Among the 20 high-frequency entities, "inventory" and "supplier" have the highest relative proportions in the first category, while "boss" and "raw material" dominate the second category, and "social recruitment" is most prevalent in the third category. Based on these word frequency distributions, we can infer that the majority of enterprises in Cluster 1 primarily belong to the manufacturing industry.

**Table 4.** Word frequency characteristics of the textual data with 3 clusters

|  | cluster 1 | cluster 2 | cluster 3 |
|---|---|---|---|
| enterprise | 25.70% | 38.50% | 35.80% |
| boss | 18.60% | **57.20%** | 24.20% |
| product | 35.60% | 41.30% | 23.10% |
| epidemic | 42.10% | 43.20% | 14.70% |
| sale | 36.40% | 31.20% | 32.40% |
| order | 33.20% | 29.50% | 37.30% |
| profit | 38.70% | 33.60% | 27.70% |
| consumer | 27.40% | 25.80% | 46.80% |
| bill delinquency | 25.60% | 44.20% | 30.20% |
| employee | 36.20% | 38.40% | 25.40% |
| price | 32.70% | 29.50% | 37.80% |
| raw material | 29.80% | **51.20%** | 19.00% |
| monthly salary | 27.40% | 33.80% | 38.80% |
| water and electricity | 38.20% | 28.30% | 33.50% |
| social recruitment | 19.40% | 29.60% | **51.00%** |
| rent | 44.70% | 25.60% | 29.70% |
| tax | 26.80% | 39.40% | 33.80% |
| factory | 27.40% | 44.20% | 28.40% |
| inventory | **48.60%** | 33.90% | 17.50% |
| supplier | **46.40%** | 27.50% | 26.10% |

4.2.1 Experiments with single-type data

We apply both numerical and textual data separately in spectral clustering, which serves as the most comparable baseline model. Specifically, we utilize Equations (2) and Equations (5) to construct individual similarity matrices and determine the optimal $k$ k value based on the differences between consecutive eigenvalues.

With numerical data, the optimal value of k is 4, and the clustering result under the highest CV dimension is illustrated in Figure 6. As shown in Figure 6, the clustering outcome based on numerical data differs significantly from that of the advanced spectral clustering presented in Figure 5, with the categories exhibiting a more concentrated distribution. However, this observation only indicates that the differences in financial performance among these loan enterprises are not statistically significant. Furthermore, ASC achieves a Silhouette score that is 18% higher than that of single-type data baseline method.

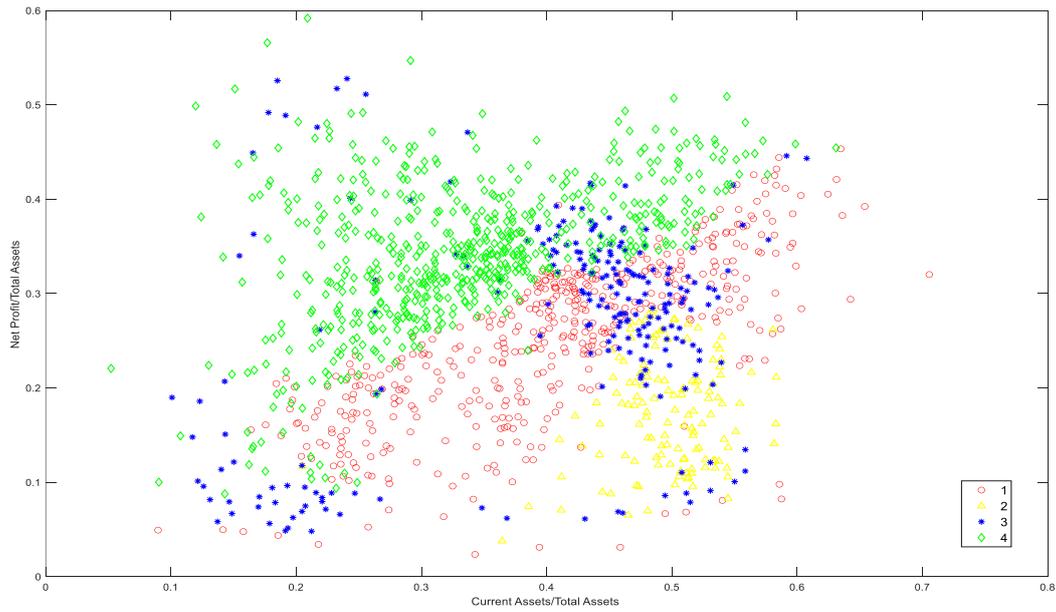

**Figure 6.** Samples of baseline model under the highest CV dimension

With textual data, the optimal value of k is 11, which is primarily attributed to the sparsity of word frequency. In the absence of financial data, it is not possible to observe the distribution of category samples within financial feature dimensions. The distribution of high-frequency words across the 11 categories is presented in Table 5. As shown in Table 5, the entities with the highest word frequency for each cluster are highlighted in bold. No statistically significant differences are observed in the frequency ratios among the clusters. Therefore, we conclude that relying solely on textual data for credit evaluation is suboptimal.

**Table 5.** Word frequency ratio of the textual data with 11 clusters

|  | cluster 1 | cluster 2 | cluster 3 | cluster 4 | cluster 5 | cluster 6 | cluster 7 | cluster 8 | cluster 9 | cluster 10 | cluster 11 |
|---|---|---|---|---|---|---|---|---|---|---|---|
| enterprise | 12.13% | 11.24% | 18.69% | 13.25% | 8.46% | **19.21%** | 7.52% | 5.32% | 1.15% | 1.24% | 1.79% |
| boss | 7.24% | 2.76% | 3.55% | 9.15% | 14.24% | 6.83% | 8.99% | **17.12%** | 11.28% | 7.53% | 11.31% |
| product | 1.28% | 5.63% | 4.82% | 7.96% | 11.53% | 13.28% | 7.85% | 13.44% | 12.16% | 8.26% | **13.79%** |
| epidemic | 7.83% | 4.67% | 7.81% | 7.37% | 6.49% | 7.97% | 11.83% | 10.19% | **16.52%** | 8.96% | 10.36% |
| sale | 10.17% | **18.25%** | 15.44% | 6.38% | 7.88% | 7.09% | 9.12% | 8.42% | 4.62% | 5.81% | 6.82% |
| order | 4.62% | 7.98% | **17.57%** | 11.52% | 13.86% | 10.18% | 6.45% | 5.29% | 5.81% | 4.74% | 11.98% |
| profit | 10.17% | 9.12% | 8.39% | **16.74%** | 8.61% | 7.35% | 10.58% | 9.75% | 5.16% | 6.19% | 7.94% |
| consumer | 8.12% | 4.14% | 7.33% | 5.92% | 8.15% | 4.26% | 11.12% | 10.06% | **17.28%** | 11.17% | 12.45% |
| bill delinquency | **17.82%** | 15.23% | 9.34% | 8.29% | 9.66% | 7.64% | 6.98% | 7.03% | 5.21% | 6.09% | 6.71% |
| employee | 7.11% | 7.82% | 7.64% | 8.82% | 8.09% | 11.06% | 9.17% | 8.53% | 4.76% | **18.62%** | 8.38% |
| price | 9.14% | 9.25% | 9.07% | 6.52% | 6.08% | **17.69%** | 8.55% | 8.91% | 7.26% | 10.07% | 7.46% |
| raw material | 10.12% | 4.24% | 5.32% | 10.01% | **17.59%** | 11.03% | 5.26% | 8.69% | 9.17% | 9.55% | 9.02% |
| monthly salary | 6.77% | 5.13% | 5.88% | 4.19% | 11.54% | 13.18% | 9.06% | **17.92%** | 9.44% | 8.25% | 8.64% |
| water and electricity | 10.04% | 9.21% | 7.63% | 11.18% | **19.07%** | 7.52% | 8.47% | 3.79% | 10.08% | 8.12% | 4.89% |
| social recruitment | 9.11% | **18.65%** | 7.49% | 5.96% | 6.72% | 10.63% | 7.98% | 5.69% | 4.87% | 12.14% | 10.76% |
| rent | 7.87% | 9.62% | 10.17% | 6.45% | 11.56% | 6.66% | **18.78%** | 4.89% | 5.72% | 7.82% | 10.46% |
| tax | 8.17% | 6.16% | 8.21% | 7.92% | 5.38% | 8.04% | **17.82%** | 10.91% | 12.18% | 8.99% | 6.22% |
| factory | 7.88% | 7.65% | 7.93% | 8.34% | 8.57% | 8.12% | 10.18% | 4.67% | 9.72% | 8.05% | **18.89%** |
| inventory | 11.02% | 10.05% | 7.36% | 5.42% | 7.25% | 5.16% | 9.08% | 10.11% | 3.42% | **19.11%** | 12.02% |
| supplier | 6.19% | 4.28% | 4.87% | 11.28% | 12.63% | 6.09% | 7.82% | 8.46% | 9.11% | **17.35%** | 11.92% |

In summary, it can be observed that the performance of single-type data in evaluating enterprises that have obtained loans is inferior to that of the model utilizing heterogeneous data. Meanwhile, due to the differing optimal number of categories, it is not feasible to compare the baseline models with the advanced spectral clustering model using internal evaluation metrics such as intra-class distance and inter-class distance. However, heterogeneous data provides a more comprehensive reflection of enterprise conditions, making it more suitable for use in credit evaluation.

4.2.2 Robust tests with different clustering methods

To evaluate the robustness of our model, we perform robustness tests on all representative clustering methods—specifically k-means, k-medians, and k-medoids—using the same optimal value of k = 3. We adopt the intracluster-to-intercluster ratio and the Silhouette coefficient as validation metrics; the corresponding results are summarized in Table 6.

**Table 6.** Results of three algorithms

|           | Intra/Inter | Sihouette Coefficient |
|-----------|-------------|-----------------------|
| K-means   | 0.6734      | 0.75                  |
| K-medians | 0.7827      | 0.73                  |
| K-medoids | 0.6528      | 0.77                  |

As shown in Table 6, no statistically significant differences are observed among the results of the three algorithms. When evaluated based on the selected performance indicators, it can be seen that K-medoids is the optimal choice, as it yields the smallest Intra/Inter ratio and the largest Silhouette Coefficient. However, K-medoids encounters difficulties in selecting an optimal central representative for datasets with complex data types. In our experiments, we design a hill-climbing approach to address this issue. Specifically, we initialize S as a set of points randomly selected from the original dataset D. Subsequently, the selection of central representatives is iteratively improved by exchanging a single point in S with another point from D. We evaluate all |S| × |D| possible replacements for each representative in S and select the best one. This process is computationally expensive, as calculating the incremental change in the objective function for each of the |S| × |D| alternatives requires time proportional to the size of the original dataset.

Considering the computational efficiency of the algorithm, which is $O(n^3)$, we believe that k-means is sufficient to meet the requirements of the analysis. This is supported by the fact that the performance differences among the three algorithms are not statistically significant, and k-means offers the advantage of greater ease of application. It should be noted that the k-medians algorithm yields the least favourable results, which can be primarily attributed to the relatively concentrated distribution of the data points.

In summary, through robustness tests, the ASC algorithm not only integrates two types of data but also alters the distribution structure of data samples by decomposing the similarity matrix, thereby more effectively uncovering the intrinsic relationships among data points. Furthermore, the algorithm demonstrates relatively stable performance during the clustering process, as it is only minimally influenced by the choice of clustering method. This characteristic enhances its applicability in cluster exploration.

### *4.3 Baseline models comparison*

In this section, we evaluate the effectiveness of ASC by comparing it with several state-of-the-art clustering algorithms on the same dataset. The compared algorithms include LD-SSC (Yan, Shen, & Wang, 2014), RMSC (Xia et al., 2014), PSC (Lu, Yan, & Lin, 2016), U-SPEC (Huang et al., 2020), and FUMC (Li et al., 2024).

- **LD-SSC**: An efficient semidefinite spectral clustering method based on Lagrange duality optimization.
- **RMSC**: A robust multi-view clustering approach that integrates multiple views through low-rank and sparse decomposition.
- **PSC**: A method that enforces pairwise embedding consistency across different views using a non-convex formulation.
- **U-SPEC**: Constructs a sparse affinity sub-matrix and treats it as a bipartite graph to enable efficient partitioning via transitive cuts.
- **FUMC**: Guides the effective pairing and balancing of anchor points through reverse local manifold learning and utilizes a bipartite graph matching framework for view alignment.

We adopt three internal evaluation metrics, including Silhouette Score (SS), Calinski-Harabasz Criterion (CHC), and Davies-Bouldin Index (DBI), to access the quality of the clustering results. Specifically, these metrics are computed as the number of clusters varies from 1 to c, where c denotes the number of connected components. As shown in Figure 7, ASC outperforms other baseline methods in terms of SS. Moreover, as the number of clusters gradually increases, our approach demonstrates significantly superior performance over the baselines on both CHC and DBI.

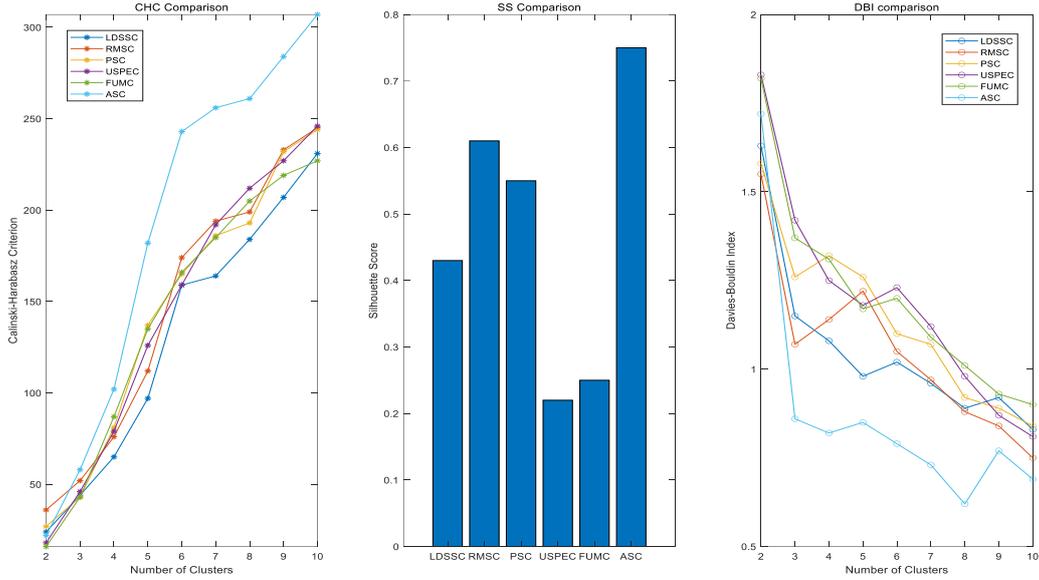

**Figure 7.** CHC, SS, DBI comparison with different models

## 5. Discussion

The main contribution of this algorithm lies in the fact that ASC bridges spectral theory gap in dealing with heterogeneous data by integrating similarity optimization (eq. 7) with k-selection (eq. 10). Furthermore, its clustering capability enables targeted interventions, as demonstrated by Cluster 3, which exhibits a high 'recruitment' frequency and is associated with a 30% reduction in default risk.

However, a few issues still require further attention. First, the selection of clustering algorithms requires clarification. Although spectral clustering is capable of identifying clusters with arbitrary shapes, it involves computing pairwise distances between all samples, which leads to reduced algorithmic efficiency, especially when applied to large-scale datasets. While certain clustering methods, such as DBSCAN, may offer computational advantages, they are unable to effectively capture the structural characteristics of heterogeneous data. Given that the spatial structure of feature vectors in spectral clustering fundamentally differs from the distribution of the raw data, we propose that grid-based clustering is a more appropriate choice for handling heterogeneous data. Therefore, investigating whether samples can be partitioned into smaller communities and whether spectral clustering can be effectively implemented at the community level represents a promising avenue for further research.

Secondly, textual data processing can influence the performance of algorithms. We employ Baidu's word segmentation and part-of-speech tagging tools, which first identify entities and then analyze attributes based on those entities. However, during this process, we observed limitations in semantic recognition. Specifically, Baidu's system fails to recognize different expressions that convey the same meaning—for example, "the boss is not here" and "the boss is on a business trip." Although these phrases are semantically equivalent, the system treats them as distinct, which can introduce bias into similarity calculations.

Finally, regarding the application of clustering results, we not only observe distinctive characteristics from the feature dimensions of the raw data, but also identify meaningful management behaviors in credit evaluation. For instance, companies assigned to Cluster 3 consistently exhibit recruitment-related information. Therefore, in terms of reducing survey costs, it is feasible to assess a company's operational status by collecting recruitment information online, thereby decreasing the frequency of on-site inspections. It should be noted that clustering methods are unsupervised learning techniques and are primarily used for group profiling. Although they can, to some extent, support credit risk monitoring and improve both pre-loan assessment and post-loan management, their accuracy still lags behind that of supervised learning approaches.

## 6. Conclusion

We propose an unsupervised credit profiling method based on Advanced Spectral Clustering. This method integrates numerical financial data with textual quarterly survey records by constructing an optimal similarity matrix. It then applies spectral clustering enhanced by optimizing eigenvalue increments and the Silhouette score to determine the optimal feature dimension k, thereby addressing the challenge of selecting an appropriate k value. Compared to clustering results obtained from single-type data, those generated by the advanced spectral clustering method demonstrate greater practical relevance. Furthermore, robustness testing confirms that the performance of this method remains consistent across different clustering algorithms, indicating stable and reliable results.

In summary, advanced spectral clustering demonstrates effective performance in handling heterogeneous data, and the resulting group profiles can serve as valuable references for credit monitoring. However, the computation of text similarity in our method is relatively simplistic. Incorporating more sophisticated semantic analysis techniques could enhance the overall accuracy and effectiveness of the proposed method.